%% file: nns.tex
\PassOptionsToPackage{table}{xcolor}

\documentclass{article}
\usepackage[utf8]{inputenc}
\usepackage[T1]{fontenc}
\usepackage{microtype}
\usepackage{graphicx}
\usepackage{subfigure}
\usepackage{booktabs}
\usepackage{comment}
\usepackage{arydshln}
\usepackage{makecell}
\usepackage{hyperref}
\usepackage{mathtools}
\usepackage{color}
\usepackage{xcolor}
\usepackage{listings}
\usepackage{CJKutf8}
\usepackage{url}
\usepackage{multicol,multirow}
\usepackage{algorithmic}
\usepackage{algorithm}
\usepackage{textcomp}
\usepackage{bm}
\usepackage{parskip}
\usepackage{amsmath,amsthm,amssymb,amsfonts}
\usepackage{times}
\usepackage{latexsym}
\usepackage{tabularx}
\usepackage{verbatim}
\usepackage{fancyvrb}
\usepackage{enumitem}
\usepackage{geometry}
\usepackage{float}
\usepackage{caption}
\usepackage[most]{tcolorbox}
\usepackage{longtable}
\usepackage{adjustbox}
\usepackage{etoolbox}
\usepackage{fontawesome5}
\usepackage[normalem]{ulem}
\usepackage{nicematrix}
\usepackage{pgfplots}
\usepackage{array}
\usepackage{marvosym}

\renewcommand{\arraystretch}{0.8}
\setlist{nolistsep}
\tcbuselibrary{listings,skins,breakable}

\definecolor{codegreen}{HTML}{189399}
\definecolor{promptblue}{RGB}{200,230,255}
\definecolor{exemplargreen}{RGB}{220,255,220}
\definecolor{protocolyellow}{RGB}{255,255,200}
\definecolor{taskblue}{RGB}{0,99,177}
\definecolor{refgreen}{RGB}{0,150,85}
\definecolor{subviolet}{RGB}{131,76,190}
\definecolor{templateblue}{RGB}{1,128,134}
\definecolor{refgreenDark}{RGB}{0,90,45}
\definecolor{analysisblue}{HTML}{1E90FF}
\definecolor{algoyellow}{HTML}{A0522D}
\definecolor{kwblue}{HTML}{005CFF}
\definecolor{strred}{HTML}{B80034}
\definecolor{codebg}{RGB}{245,248,250}
\definecolor{argsc}{RGB}{0,128,128}
\definecolor{code-highlight-blue}{HTML}{2696f0}
\definecolor{code-highlight-green}{HTML}{7eb547}
\definecolor{code-highlight-yellow}{HTML}{fdcc3b}
\definecolor{code-highlight-purple}{HTML}{ab4abb}
\definecolor{code-highlight-red}{HTML}{f3473a}
\colorlet{subvioletstrong}{subviolet!80!black}

\lstdefinestyle{codestyle}{
    language=Python,
    basicstyle=\footnotesize\ttfamily,
    frame=single,
    numbers=left,
    numberstyle=\tiny,
    xleftmargin=1.5em,
    framexleftmargin=1.5em,
    keywordstyle=\color{taskblue},
    commentstyle=\itshape\color{gray},
    stringstyle=\color{orange},
    showstringspaces=false,
    breaklines=true,
    tabsize=2
}
\lstdefinestyle{py}{
    language=Python,
    basicstyle=\tiny\ttfamily,
    numbers=left,
    numberstyle=\tiny\color{gray},
    stepnumber=1,
    keywordstyle=\color{kwblue}\bfseries,
    commentstyle=\color{gray},
    stringstyle=\color{strred},
    breaklines=true,
    showstringspaces=false,
    tabsize=4,
    backgroundcolor=\color{templateblue!5},
    frame=single,
    xleftmargin=3em,
    framexleftmargin=2.5em,
    rulecolor=\color{codebg},
    framesep=6pt,
    literate={Args}{{\textcolor{argsc}{Args}}}4
}
\lstdefinestyle{afteroptimcode}{
    language=Python, numbers=left, frame=none,
    numberstyle=\tiny,
    numbersep=3pt,
    xleftmargin=2pt,
    framexleftmargin=0pt,
    basicstyle=\ttfamily\footnotesize,
    keywordstyle=\color{subviolet},
    breaklines=true,
    columns=fullflexible
}
\lstdefinestyle{beforeoptimcode}{
    language=Python, numbers=left, frame=none,
    numberstyle=\tiny,
    numbersep=3pt,
    xleftmargin=2pt,
    framexleftmargin=0pt,
    basicstyle=\ttfamily\footnotesize,
    keywordstyle=\color{subviolet},
    breaklines=true,
    columns=fullflexible
}

\newcommand{\emailfootnote}[1]{%
    \textsuperscript{\large\Letter}%
    \begingroup
    \renewcommand{\thefootnote}{\large\Letter}%
    \footnotetext{#1}%
    \endgroup
}

\title{CRINN: Contrastive Reinforcement Learning for Approximate\\ Nearest Neighbor Search}
\author{Xiaoya Li, Albert Wang, Guoyin Wang, Chris Shum and Jiwei Li}
\date{\textbf{\large Ornith Team}\\[0.15cm]
    \includegraphics[scale=0.05]{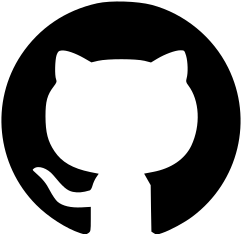}
    \href{https://github.com/ornith-ai/CRINN}{{\large github.com/ornith-ai/crinn}}
}

\begin{document}

\maketitle

\begin{abstract}
Approximate nearest-neighbor search (ANNS) algorithms have become increasingly critical for recent AI applications, particularly in retrieval-augmented generation (RAG) and agent-based LLM applications. In this paper, we present CRINN, a new paradigm for ANNS algorithms.
CRINN treats ANNS optimization as a reinforcement learning problem where execution speed serves as the reward signal.
This approach enables the automatic generation of progressively faster ANNS implementations while maintaining accuracy constraints.
Our experimental evaluation demonstrates CRINN's effectiveness across six widely-used NNS benchmark datasets. When compared against state-of-the-art open-source ANNS algorithms, CRINN achieves best performance on three of them (GIST-960-Euclidean, MNIST-784-Euclidean, and GloVe-25-angular), and tied for first place on two of them (SIFT-128-Euclidean and GloVe-25-angular). The implications of CRINN's success reach well beyond ANNS optimization: It validates that LLMs augmented with reinforcement learning can function as an effective tool for automating sophisticated algorithmic optimizations that demand specialized knowledge and labor-intensive manual refinement.
\emailfootnote{Email: \{xiaoya\_li, albert\_wang, chris\_shum, jiwei\_li\}@ornith.ai}
\end{abstract}

\vspace{-0.1cm}
\input{./images/overall_results_plot}

\section{Introduction}
Approximate nearest-neighbor search (ANNS) \cite{malkov2018efficient,manohar2024parlayann,douze2024faiss,fu2017fast,lin2021pyserini,jayaram2019diskann,dong2011efficient} aims at finding data points that are closest to a query point in high-dimensional spaces while trading off a small amount of accuracy for significant speedup over exact search methods. ANNS is of growing importance due to the unprecedented popularity of retrieval-augmented generation (RAG)~\cite{Lewis2020RetrievalAugmentedGF,Guu2020REALMRL,Meng2021FastNN} and agent-based LLM applications~\cite{Park2023GenerativeAI}, which require fast retrieval of relevant information from massive vector databases.
Existing widely-used open-source projects use fundamental algorithms such as Vamana \cite{jayaram2019diskann} and HNSW \cite{malkov2018efficient} as the backbone: DiskANN and ParlayANN \cite{manohar2024parlayann} build upon Vamana, while FAISS \cite{douze2024faiss}, Vespa\footnote{\url{https://docs.vespa.ai/en/nearest-neighbor-search.html}}, Weaviate\footnote{\url{https://weaviate.io/blog/vector-search-explained}}, Qdrant\footnote{\url{https://github.com/qdrant/qdrant}}, Milvus\footnote{\url{https://github.com/milvus-io/milvus}}, and GLASS~\cite{PyGlass}\footnote{\url{https://github.com/hhy3/pyglass}} leverage HNSW, proposing different levels of optimization to cater for various scenarios.

Existing optimizations for ANNS are mostly manual, where humans identify bottlenecks through profiling tools, analyze cache miss patterns, hand-tune parameters like graph degree and search beam width, carefully design data structures for memory locality, and iteratively experiment with different algorithmic variants based on hardware characteristics.
This process requires deep expertise in computer architecture, parallel programming, and the mathematical properties of ANNS algorithms.
With the increasing power of LLMs in code generation~\cite{Guo2024DeepSeekCoderWT,Hui2024Qwen25CoderTR}, a natural question arises: can LLMs facilitate optimization by automatically generating and testing algorithmic improvements, learning from the execution speeds of previous implementations to propose better solutions, and discovering novel optimization patterns that human engineers might overlook?

In this paper, we propose CRINN, a reinforcement learning-augmented LLM framework for automated ANNS optimization. The core of CRINN is a contrastive RL model that performs comparative analysis of previously generated code variants alongside their execution metrics, enabling the model to learn by distinguishing between effective and ineffective optimization strategies and generate better solutions.
Through this contrastive learning approach, CRINN develops an understanding of which code patterns lead to performance improvements and which cause degradation. The generated code variants are evaluated based on execution time, with faster implementations receiving higher rewards in the RL training process. This reward signal drives the LLM to iteratively generate progressively more efficient ANNS implementations. By learning from the performance outcomes of its own generated code, CRINN effectively transforms the manual optimization process into an automated search through the space of possible implementations.

Our experimental evaluation demonstrates CRINN's effectiveness across six widely-used NNS benchmark datasets~\cite{Aumller2018ANNBenchmarksAB}. When compared against state-of-the-art open-source ANNS algorithms, CRINN achieves best performance on three of them (GIST-960-Euclidean, MNIST-784-Euclidean, and GloVe-25-angular); tied for first place on two of them (SIFT-128-Euclidean and GloVe-25-angular).
More importantly, the success of CRINN carries broader implications beyond ANNS optimization. It demonstrates that RL-augmented LLMs can serve as powerful tools for automating complex algorithmic optimizations that traditionally require deep domain expertise and extensive manual tuning. As the demand for efficient vector search continues to grow with the proliferation of RAG and agent-based LLM applications, automated optimization frameworks like CRINN will become increasingly valuable for maintaining competitive performance across evolving hardware architectures and application requirements.

\section{Background: HNSW}
\label{background}
We use HNSW (Hierarchical Navigable Small World) \cite{malkov2018efficient} as the optimization backbone. HNSW is a state-of-the-art graph-based ANNS algorithm that achieves high recall rates with logarithmic search complexity. Here we describe the core modules that most HNSW-based implementations adopt, on which our RL optimization is performed.

\subsection{Graph Construction}
In HNSW, we first need to build a graph for base vectors, where each vector is represented as a node in the graph. As the search is performed through this graph for a given query, graph construction is a key step in HNSW.

The graph construction module builds the graph through incremental insertion of vectors. HNSW constructs a multi-layer graph where each element is assigned to multiple layers based on an exponentially decaying probability distribution. Each new base vector is assigned to layer 0 with probability 1, and to each subsequent layer $l$ with probability $p^l$, where $p$ is typically set to $1/2\ln(2)$. This creates a hierarchical structure similar to skip lists, enabling efficient navigation during search.

For each layer where a vector is present, the algorithm performs a greedy search from the layer's entry point to find the $M$ nearest neighbors. The search parameter $ef$ controls the number of candidates explored during this neighbor selection process. The algorithm maintains $M$ connections for upper layers and $M \times 2$ connections for the bottom layer (layer 0) to ensure rich connectivity. It is worth noting that $ef$ is a crucial parameter in HNSW, controlling the tradeoff between recall and speed. Larger values of $ef$ mean exploring more candidates during the search phase, thus achieving higher recall at the cost of reduced speed.
HNSW employs a heuristic pruning strategy to optimize the graph structure. This strategy maintains connectivity while promoting the small-world property by prioritizing diverse connections over purely nearest neighbors. Redundant edges that don't contribute to search efficiency are removed, resulting in a graph that balances accuracy with traversal efficiency.

\subsection{Search}
Given a constructed graph and an input query, the search module performs k-nearest neighbor search through a two-phase hierarchical process. The search begins with the upper layer, starting from the global entry point at the top layer. The algorithm greedily traverses each layer to find the single nearest neighbor, using this neighbor as the entry point for the next lower layer. This process continues until reaching the bottom layer (layer 0), effectively narrowing down the search space.

At layer 0, the algorithm switches to a more exhaustive exploration strategy. It initializes a candidate set with the entry point from layer 1 and maintains a priority queue of $ef$ nearest candidates (where $ef \geq k$). The search iteratively explores neighbors of the closest unexplored candidate, updating the result set whenever closer neighbors are discovered. The process terminates when no remaining candidate can improve the current results, ensuring that the $k$ nearest neighbors have been found with high probability.

\subsection{Refinement}
Most HNSW-based algorithms incorporate refinement modules that enhance search efficiency through various optimization strategies. One common approach is quantized preliminary search, where vectors are compressed to int4 or int8 representations for rapid similarity estimation. Product quantization further improves this by dividing vectors into subspaces and quantizing each independently. Asymmetric distance computation keeps queries in full precision while using quantized database vectors, balancing speed and accuracy.
Hierarchical pruning strategies form another class of refinements. These include layer-wise filtering that uses coarse distance estimates to prune entire graph regions early in the search process. Batch processing amortizes memory access costs by handling multiple queries simultaneously, while adaptive beam search dynamically adjusts the search width based on query difficulty.

\section{Improving ANNS with Contrastive Reinforcement Learning}
In this section, we describe how we optimize an ANNS algorithm using contrastive RL \cite{li2025cuda} in detail.

\subsection{Overview}
An ANNS algorithm usually contains multiple modules, e.g., graph construction, search, refinement as described in Section~\ref{background}.
We treat each module in ANNS as independent and optimize module by module sequentially using contrastive RL.

The core idea of contrastive RL is to incorporate previously evaluated code implementations and their execution times directly into the optimization prompt. By comparing the performance of different implementations, the model identifies code patterns associated with higher execution efficiency and proposes improved solutions.

Each generated implementation is evaluated using its execution performance, which provides the reward signal for reinforcement learning. The resulting feedback is used to update the model parameters, creating an iterative optimization process that progressively improves the efficiency of ANNS implementations.

\input{./tables/rl_prompt_template}
\subsection{Prompt Construction}

Here we describe the construction of prompts provided to the LLM during contrastive-RL training. The prompt comprises the following structured components:

\begin{enumerate}
    \item \textbf{Task Description:} A detailed description of the current ANNS module to optimize, including input/output requirements, performance metrics, and optimization objectives with emphasis on execution speed.
    \item \textbf{Previous Implementations with Speed:} Previously generated ANNS implementations paired with their scores indicating execution speed.
    \item \textbf{Generation Protocol:} Explicit instructions defining the required output format.
    \item \textbf{Critical Requirements:} Explicit instructions for the required output content.
\end{enumerate}

The model's response must contain three structured components:
\begin{enumerate}
    \item \textbf{Performance Analysis:} A comparative analysis identifying which previous ANNS module implementations achieved superior speed performance and the underlying algorithmic or implementation factors contributing to their fast execution.
    \item \textbf{Algorithm Design:} A high-level description of the proposed optimization strategy, outlining key techniques and improvements for accelerating execution speed as numbered points in natural language.
    \item \textbf{Code Implementation:} The detailed code implementation.
\end{enumerate}

A detailed example of the prompt structure is shown in Table~\ref{tab:prompt_example}.
For selecting historical implementations, we maintain a performance-indexed database of code candidates produced and evaluated during previous optimization iterations.
Implementations are sampled according to their measured execution performance using a temperature-scaled softmax distribution:
\begin{equation}
P(B_i) = \frac{\exp\left( (s_i - \mu)/\tau \right)}{\sum_j \exp\left((s_j - \mu)/\tau\right)}
\label{eq:bucket_sampling}
\end{equation}
where $s_i$ denotes the score of a code sample in the database, $\mu$ denotes the mean score across all codes in the database, and $\tau$ is the temperature parameter governing the exploration-exploitation tradeoff.

\subsection{Speed Reward}
Giving a generated code implementation a proper reward, which must be a \textbf{scalar}, is crucial in training our system. The reward serves two important purposes: (1) guiding parameter updates in reinforcement learning and (2) constructing prompts for contrastive analysis.

Unfortunately, constructing a reward that effectively captures general code speed performance in ANNS is not as straightforward as it seems.
In the ANNS setting, performance is characterized by two interdependent metrics: queries per second (QPS) and recall. The only parameter we can adjust is $ef$, which controls the number of neighbors explored during search: higher values of $ef$ lead to lower QPS and higher recall.
However, this creates a comparison problem: two implementations with identical $ef$ can exhibit different combinations of QPS and recall. Simply using QPS as the reward would be unfair without accounting for recall differences.

This challenge explains why the ANNS literature rarely reports single quantitative metrics (unlike traditional ML metrics such as accuracy or F1 score). Instead, researchers typically present QPS-recall curves (as in Figure~\ref{fig:overall_result_plot}) that visualize the entire performance tradeoff space.
A seemingly straightforward solution would be to fix one metric and optimize the other---either fix recall and maximize QPS, or fix QPS and maximize recall. Unfortunately, this approach is infeasible because $ef$ takes discrete values. We cannot continuously tune $ef$ to achieve specific target values of QPS or recall; instead, we can only obtain a discrete set of (QPS, recall) points corresponding to different $ef$ settings.
This discretization prevents us from making fair comparisons at fixed performance levels.
This discrete nature creates other evaluation challenges: some algorithms may excel in low-recall regions but perform poorly in high-recall regions, while others show the opposite pattern; some algorithms may not even produce data points in certain regions of the spectrum. For example, certain high-quality algorithms often cannot achieve low recall values regardless of $ef$ values.
This incomplete coverage of the performance spectrum adds significant complexity to reward estimation, where the RL framework requires the reward to be a single scalar value.

To address these challenges, we employ the following evaluation strategy: given a module implementation, we first vary $ef$ across different values to obtain a comprehensive set of (QPS, recall) points. We then filter these points to retain only those within the recall range of [0.85, 0.95] and compute the area under the curve formed by these points. This area serves as our scalar reward.

We choose the [0.85, 0.95] recall range for several reasons. First, we primarily care about algorithms that achieve reasonable recall levels---performance at very low recall is generally not useful for practical applications. Second, in the high-recall region above 0.95, data points become increasingly sparse, and some algorithms may not even produce points in this range, leading to unstable reward estimation. The [0.85, 0.95] range thus represents a sweet spot where most algorithms have sufficient data points and where performance differences are most meaningful for real-world deployment.

\subsection{RL Training}

For reinforcement learning training, we employ the Group Relative Policy Optimization (GRPO) approach \cite{shao2024deepseekmath}. For each input prompt $q$ augmented with previously evaluated code implementations and their execution metrics, we generate $G$ code completions from the current policy $\pi_{\mathrm{old}}$, represented as $\{d_1,d_2,\ldots,d_G\}$. The corresponding reward scores are denoted by $\mathbf{r}=(r_1,r_2,\ldots,r_G)$.
To ensure training stability, rewards undergo smoothing following \cite{li2025cuda}. Following the GRPO methodology, we normalize rewards within each group:
\begin{equation}
\hat{r}_i=\frac{r_i-\operatorname{mean}(\mathbf{r})}{\operatorname{std}(\mathbf{r})}.
\end{equation}

The GRPO training objective maximizes the following function:
\begin{align}
\mathcal{L}_{\mathrm{GRPO}}(\theta)
&=\mathbb{E}_{\substack{q\sim P(q)\\\{d_i\}_{i=1}^{G}\sim\pi_{\theta_{\mathrm{old}}}(\cdot\mid q)}}
\Bigg[\frac{1}{G}\sum_{i=1}^{G}\frac{1}{|d_i|}\sum_{t=1}^{|d_i|}
\Bigg(\min\Bigg(
\frac{\pi_\theta(d_{i,t}\mid q,d_{i,<t})}{\pi_{\theta_{\mathrm{old}}}(d_{i,t}\mid q,d_{i,<t})}\hat{r}_i, \nonumber\\
&\qquad\qquad\quad
\operatorname{clip}\!\left(
\frac{\pi_\theta(d_{i,t}\mid q,d_{i,<t})}{\pi_{\theta_{\mathrm{old}}}(d_{i,t}\mid q,d_{i,<t})},
1-\varepsilon,1+\varepsilon
\right)\hat{r}_i\Bigg)
-\beta D_{\mathrm{KL}}[\pi_\theta\Vert\pi_{\mathrm{ref}}]\Bigg)\Bigg].
\end{align}

Here:
\begin{itemize}
    \item $\pi_\theta$ represents the policy network under optimization.
    \item $\pi_{\theta_{\mathrm{old}}}$ denotes the policy from the preceding training step.
    \item $\varepsilon$ controls the clipping range for policy updates.
    \item $\beta$ is a regularization parameter balancing exploration and adherence to the reference policy.
    \item $D_{\mathrm{KL}}$ measures the Kullback--Leibler divergence between current and reference distributions.
\end{itemize}
For comprehensive details on GRPO, we direct readers to \cite{shao2024deepseekmath}.
The model's parameters are updated through this GRPO objective, utilizing contrastive prompts enriched with comparative implementations and their measured performance.

\subsection{Using Glass as the RL Starting Point}
Instead of implementing an ANNS algorithm from scratch, we use an existing open-source ANNS algorithm as a starting point and progressively optimize its constituent modules. We selected GLASS as our initial baseline due to its balanced performance across diverse evaluation datasets. GLASS is a recent graph-based ANNS algorithm that achieves efficient graph construction while maintaining competitive search performance.
It is important to note that CRINN is a general optimization framework---while we demonstrate its effectiveness using GLASS, it can take any existing open-source ANNS algorithm as a starting point and progressively evolve its implementation for improved performance.
The choice of GLASS simply provides a strong foundation for showcasing CRINN's capabilities across different optimization scenarios. We will update the results based on another strong baseline ParlayANN in the upcoming version of this project.
Using GLASS, we sequentially optimize the code across the three key modules in HNSW: \textbf{graph construction}, \textbf{search}, and \textbf{refinement}.

\section{Experiments and Analysis}
\subsection{Datasets and Baselines}

We use the following six widely used datasets for evaluation: SIFT-128-Euclidean, GIST-960-Euclidean, MNIST-784-Euclidean, GloVe-100-Angular, GloVe-25-Angular, and NYTimes-256-Angular.
SIFT-128-Euclidean consists of 128-dimensional SIFT features extracted from images. GIST-960-Euclidean contains 960-dimensional GIST global descriptors for images. MNIST-784-Euclidean is composed of 784-dimensional vectors representing flattened 28$\times$28 pixel images of handwritten digits. GloVe-100-Angular and GloVe-25-Angular are datasets of word embeddings trained by the GloVe algorithm on a large text corpus, with dimensions 100 and 25, respectively. NYTimes-256-Angular contains 256-dimensional bag-of-words vectors from New York Times articles. The statistics for these datasets are summarized in Table~\ref{tab:datasets}.

\begin{table}
\centering
\setlength{\tabcolsep}{5pt}
\renewcommand\arraystretch{1.1}
\begin{tabular}{@{\hspace{3pt}}l@{\hspace{12pt}}c@{\hspace{12pt}}c@{\hspace{15pt}}c@{\hspace{15pt}}c@{\hspace{3pt}}}
\toprule
\textbf{Dataset} & \textbf{D} & \textbf{LID} & \textbf{Base Vectors} & \textbf{Query Vectors} \\
\midrule
\multicolumn{5}{c}{\textit{Euclidean Distance}} \\[2pt]
SIFT-128    & 128 & 9.3  & 1,000,000 & 10,000 \\
GIST-960    & 960 & 20.5 & 1,000,000 & 1,000  \\
MNIST-784   & 784 & 14.1 & 60,000    & 10,000 \\[4pt]
\multicolumn{5}{c}{\textit{Angular Distance}} \\[2pt]
GloVe-25    & 25  & 9.9  & 1,183,514 & 10,000 \\
GloVe-100   & 100 & 12.3 & 1,183,514 & 10,000 \\
NYTimes-256 & 256 & 12.5 & 290,000   & 10,000 \\
\bottomrule
\end{tabular}
\caption{Statistics of the benchmark datasets. D is the vector dimension, and LID is the average Local Intrinsic Dimension.}
\label{tab:datasets}
\end{table}

To maintain a single generalizable codebase, we train our reinforcement learning model exclusively based on rewards from the SIFT-128 dataset. The resulting implementation is then evaluated across all datasets without modification.
It is worth noting that SIFT-128 uses Euclidean distance, which means our RL model is trained only on Euclidean rewards. This might cause problems for angular-distance datasets. However, as we will show in the experiments, the code trained on Euclidean distance generalizes well to angular-distance datasets, demonstrating the strong generalization capability of CRINN. Incorporating both Euclidean and angular distances as training rewards constitutes future work.

We employ the following widely used open-source projects as baselines:
\begin{itemize}
\item \textbf{Glass}~\cite{PyGlass}: A graph-based approximate nearest neighbor search library developed by Zilliz, utilizing hierarchical navigable small world (HNSW) graphs with optimizations for high-dimensional vector search and hardware acceleration.
Glass serves as the starting point for our RL training process, making it a natural baseline for performance comparison.

\item \textbf{ParlayANN}~\cite{Manohar2023ParlayANNSA}: A parallel approximate nearest neighbor library that leverages multi-core parallelism and cache-efficient algorithms, offering implementations of graph-based search methods optimized for shared-memory multiprocessors. In the updated version of this project, we will include the experimental setup that uses ParlayANN as the reinforcement learning starting point.

\item \textbf{NNDescent} \cite{ono2023relative}: An implementation of the NN-Descent algorithm that constructs approximate k-nearest neighbor graphs through iterative local search, efficiently handling both dense and sparse data with minimal memory overhead.

\item \textbf{PyNNDescent}\footnote{\url{https://github.com/lmcinnes/pynndescent}}: A Python implementation of NN-Descent that provides fast approximate nearest neighbor queries and KNN graph construction, with support for a wide variety of distance metrics and efficient handling of high-dimensional data.

\item \textbf{Vearch}~\cite{li2019design}\footnote{\url{https://github.com/vearch/vearch}}: A distributed vector search system designed for large-scale similarity search, combining graph-based indexing with inverted file structures to support billion-scale vector retrieval in production environments.

\item \textbf{Voyager}\footnote{\url{https://github.com/spotify/voyager}}: A library developed by Spotify for approximate nearest neighbor search, implementing hierarchical navigable small world graphs with optimizations for music and recommendation system workloads.
\end{itemize}

\section{Results}
\subsection{Major Results}
Figure~\ref{fig:overall_result_plot} presents the QPS versus recall curves for different models across six datasets. CRINN outperforms all baselines on three benchmarks: GIST-960-Euclidean, MNIST-784-Euclidean, and GloVe-25-angular. The improvement is particularly substantial for MNIST-784-Euclidean.
On SIFT-128-Euclidean and GloVe-25-angular, CRINN achieves performance comparable to the best baseline---matching ParlayANN on SIFT-128 and Vearch on GloVe-25. Among the six datasets, CRINN underperforms the best baseline only on NYTimes-256.
This performance gap likely stems from the fundamental differences between distance metrics. Since the RL-optimized code was trained exclusively on SIFT-128 using Euclidean distance, it may not effectively capture the optimization patterns required for angular similarity computations. Notably, it is common for strong models to exhibit dataset-specific weaknesses. For instance, ParlayANN performs poorly on GloVe-25-angular despite its strong performance elsewhere.
When comparing CRINN with GLASS, which serves as the RL starting point, CRINN demonstrates substantial improvements. This consistent performance gain indicates that contrastive RL can robustly and progressively optimize the code.
\input{./tables/recall}

\subsection{QPS with Fixed Recall}
To give a quantitative comparison, Table~\ref{tab:performance} presents the QPS (Queries Per Second) performance of CRINN against the best-performing baselines across six benchmark datasets at various recall levels (0.9, 0.99, 0.999).
For cases where performances are absent for specific recall levels, it indicates that none of the tested methods could reach the target recall threshold.
The results demonstrate that CRINN consistently outperforms state-of-the-art methods across most configurations, with improvements ranging from modest gains of 3.09\% to substantial speedups of 85.25\%.
CRINN shows particularly strong performance on MNIST-784, achieving up to 85.25\% improvement at 0.999 recall, and on GIST-960 at high recall levels, with a 72.68\% improvement at 0.99 recall. The SIFT-128 dataset, which was used for training the RL agent, shows consistent improvements across all recall levels, with gains decreasing as recall requirements become more stringent. Among angular distance datasets, GloVe-25 exhibits significant improvements of up to 32.01\%, while GloVe-100 shows mixed results, including a slight degradation of 5.84\% at 0.95 recall.
As mentioned above, CRINN achieves poor performance on NYTimes-256, where CRINN underperforms the best baseline by 82.85\% for the 0.9 recall setup.

\subsection{Progressive Improvements for Different Modules}
In CRINN, optimization proceeds sequentially through three modules: graph construction, search, and refinement. To evaluate the individual contribution of each module, we examine the progressive performance improvements at each optimization stage.
We demonstrate these incremental gains by measuring the average QPS improvement across fixed recall values (0.90, 0.95, 0.99, and 0.999).

Table~\ref{table:progressive} presents the result.
As can be seen, CRINN demonstrates substantial gains through all its three optimization stages. The graph construction module yields the most significant individual improvements, averaging 22.11\% across all recall values and datasets, with particularly impressive results on gist-960-euclidean (58.26\%) and mnist-784-euclidean (45.85\%). The search optimization module contributes an additional 18.30\% on average, maintaining strong performance especially for high-dimensional datasets. The refinement module, showing more modest individual gains of 9.69\%.
The diminishing returns across the three stages can be attributed to two primary factors. First, the fundamental importance hierarchy of these stages naturally leads to different improvement potentials. Graph construction is the foundation that determines the entire search space structure. In contrast, the refinement stage serves a more specialized role of fine-tuning results, where the candidates are already of reasonable quality, thus offering less dramatic improvement potential.
Second, the optimization order plays a significant role---earlier stages have more optimization opportunities available, while later stages operate on an already-improved system. The graph construction stage works with the raw, unoptimized baseline, allowing techniques like adaptive search, multi-level prefetching, and multi-entry points to capture the ``low-hanging fruit'' of performance improvements. By the time the refinement stage is reached, many inefficiencies have already been addressed, leaving less room for dramatic improvement.

The only outlier is nytimes-256-angular, which shows performance degradation across all stages, suggesting that the optimization techniques may need dataset-specific tuning for certain angular distance computations. Overall, the results validate the effectiveness of the progressive optimization strategy, with five out of six datasets showing substantial cumulative improvements ranging from 16\% to 134\%.

\input{./tables/progressive}

\section{Analysis}
Here, we conduct a detailed analysis of each of the three optimization modules, examining the specific changes introduced by contrastive RL and how they contribute to improved performance.

\subsection{Graph Construction}
We identify the following three optimization strategies discovered by contrastive RL in the graph construction module.\vspace{-5mm}
\paragraph{\bf Adaptive Search with Dynamic EF Scaling}\hspace{-3mm}, which adjusts search effort based on target recall requirements, replacing the fixed ef\_construction parameter in the original code with an adaptive value. This strategy helps allocate computational resources proportionally to quality requirements.
\begin{lstlisting}[language=C++, numbers=left, frame=single, xleftmargin=2em, framexleftmargin=1.5em]
// Old: Fixed search budget
size_t ef = ef_construction;  // Always constant

// New: Adaptive search budget based on recall needs
if (target_recall > critical_threshold)
    dynamic_ef = ef_search * (1.0 + recall_excess * 14.5);
else
    dynamic_ef = ef_search;
\end{lstlisting}

\paragraph{\bf Zero-Overhead Multi-Level Prefetching}\hspace{-3mm}, which replaces fixed prefetching with an adaptive multi-level strategy that considers neighbor density and search layer.
It reduces memory access latency compared to the fixed prefetch window in the old implementation.
\begin{lstlisting}[language=C++, numbers=left, frame=single, xleftmargin=2em, framexleftmargin=2em]
// Old: Fixed prefetch window
for (int j = 0; j < min(5, size); ++j)
    computer.prefetch(neighbors[j], 1);

// New: Adaptive multi-level prefetching
prefetch_depth = min(adaptive_depth, size);  // 24-48 based on performance
for (int j = 0; j < prefetch_depth; ++j)
    computer.prefetch(neighbors[j], 3);  // L1 cache
if (high_recall_needed)
    // Additional L2 prefetch for more neighbors
\end{lstlisting}

\paragraph{\bf Multi-Entry Point Search Architecture}\hspace{-3mm}, which maintains multiple diverse entry points for parallel exploration instead of a single global entry point.
This strategy improves recall for the same computational budget by exploring diverse graph regions simultaneously.
\begin{lstlisting}[language=C++, numbers=left, frame=single, xleftmargin=2em, framexleftmargin=2em]
// Old: Single entry point
start_node = enterpoint_node;
results = search(start_node, query);

// New: Multiple diverse entry points (up to 9)
for (node : strategic_entrypoints) {
    if (distance_to_others(node) > threshold)
        entry_points.add(node);
}
// Search from multiple starting points
for (ep : entry_points)
    results.merge(search(ep, query));
\end{lstlisting}

\subsection{Search}
For search, we identify the following three optimization strategies proposed by contrastive RL.\vspace{-5mm}
\paragraph{\bf Multi-Tier Entry Point Selection}\hspace{-3mm}, which replaces single entry point initialization with a sophisticated multi-tier system that selects from primary, secondary, and tertiary entry points based on search budget. This strategy improves search quality by starting from diverse, high-quality nodes.
\begin{lstlisting}[language=C++, numbers=left, frame=single, xleftmargin=2em, framexleftmargin=1.5em]
// Old: Single entry point
initialize_search(single_entry_point)

// New: Multi-tier entry selection
add_entry(primary_entry_point)
if search_budget > threshold_1:
    add_entry(secondary_entry_point)
if search_budget > threshold_2:
    add_entry(tertiary_entry_point)
\end{lstlisting}

\paragraph{\bf Batch Processing with Adaptive Prefetching}\hspace{-3mm}, which optimizes neighbor processing by collecting edges into batches and using enhanced prefetch strategies. This reduces random memory access and improves cache utilization.
\begin{lstlisting}[language=C++, numbers=left, frame=single, xleftmargin=2em, framexleftmargin=2em]
// Old: Fixed prefetching
for i in range(prefetch_count):
    prefetch(neighbor[i])

// New: Adaptive batch prefetching
prefetch_size = prefetch_count * batch_factor
for i in range(adaptive_prefetch_size):
    prefetch(neighbor[i])
    if processing_node[j]:
        prefetch(neighbor[j + prefetch_size])  // Look ahead
\end{lstlisting}

\paragraph{\bf Intelligent Early Termination with Convergence Detection}\hspace{-3mm}, which monitors search progress and terminates early when convergence is detected, avoiding unnecessary exploration while maintaining quality.
\begin{lstlisting}[language=C++, numbers=left, frame=single, xleftmargin=2em, framexleftmargin=2em]
// Old: Explore until pool exhausted
while has_candidates():
    process_neighbor()

// New: Smart termination
no_improvement_count = 0
while has_candidates():
    improvements = process_neighbor()
    if improvements == 0:
        no_improvement_count++
        if check_convergence(no_improvement_count):
            break  // Early termination
\end{lstlisting}

\subsection{Refinement}
For the refinement module, RL proposed the following two optimization strategies.\vspace{-5mm}
\paragraph{\bf Adaptive Memory Prefetching}\hspace{-3mm}, which replaces basic hierarchical search with an intelligent prefetching system that adapts based on edge patterns and node characteristics. This strategy significantly reduces memory latency during the refinement process.
\begin{lstlisting}[language=C++, numbers=left, frame=single, xleftmargin=2em, framexleftmargin=2em]
// Old: Basic traversal without prefetching
for each edge v in node_edges:
    if distance(v) < best_distance:
        update best_node

// New: Adaptive prefetching with lookahead
if should_prefetch:
    prefetch(edges[0])
for i, edge v in node_edges:
    prefetch(edges[i + lookahead])  // Prefetch future edges
    if distance(v) < best_distance:
        update best_node
\end{lstlisting}

\paragraph{\bf Pre-computed Edge Metadata with Pattern Recognition}\hspace{-3mm}, which enhances the refiner by pre-computing and storing edge counts for each node level. This eliminates redundant computations and enables pattern-based optimizations during refinement.
\begin{lstlisting}[language=C++, numbers=left, frame=single, xleftmargin=2em, framexleftmargin=2em]
// Old: Runtime edge counting
count = 0
for each edge in node:
    if edge != -1:
        count++

// New: Pre-computed metadata access
metadata = get_precomputed_metadata(level, node)
edge_count = metadata.count
pattern_score = metadata.intelligence_score
// Use metadata for optimization decisions
if pattern_score > threshold:
    apply_pattern_optimization()
\end{lstlisting}

\section{Related Work}
The past year has witnessed a surge of interest in leveraging LLMs and RL-augmented LLM models for code optimization. This includes significant advances in compiler optimization \cite{cummins2025llm}, assembly code optimization \cite{wei2025improving}, and CUDA kernel optimization \cite{lange2025ai,chen2025cuda,li2025cuda}. Reinforcement learning frameworks such as CodeRL~\cite{Le2022CodeRLMC} and PPOCoder~\cite{Shojaee2023ExecutionbasedCG} have emerged as powerful tools for enhancing LLM performance in code generation and optimization tasks. Notably, RLEF \cite{gehring2024rlef} demonstrates that end-to-end reinforcement learning can effectively train models to utilize execution feedback during code synthesis, achieving state-of-the-art performance on competitive programming benchmarks.

In assembly code optimization, recent breakthroughs show that PPO-trained LLMs can achieve remarkable results---reaching 96.0\% test pass rates and delivering 1.47$\times$ speedups compared to the gcc -O3 baseline \cite{wei2025improving}. Similarly, Meta's LLM Compiler \cite{cummins2024meta} achieves 77\% of the optimization potential of exhaustive autotuning searches, validating the effectiveness of LLMs for optimizing compiler intermediate representations.
For GPU-accelerated computing, CUrator \cite{lee2025curator} introduces an efficient LLM execution engine that seamlessly integrates CUDA libraries such as cuBLAS and CUTLASS, optimizing performance for modern language models. Complementing this, ComputeEval provides an open-source benchmark framework specifically designed to evaluate LLM capabilities in CUDA programming tasks, establishing standardized metrics for this emerging field.

\section{Conclusion}

In this paper, we presented CRINN, a novel framework that employs contrastive reinforcement learning-augmented LLMs to automatically optimize approximate nearest-neighbor search algorithms. By treating ANNS optimization as a reinforcement learning problem where execution speed serves as the reward signal, CRINN successfully transforms the traditionally manual and expertise-intensive optimization process into an automated search through the space of possible implementations.
Our experimental results demonstrate CRINN's effectiveness across diverse benchmark datasets, achieving best-in-class performance on three out of six benchmarks and matching state-of-the-art results on two others.
The success of CRINN carries broader implications beyond ANNS optimization. It demonstrates that RL-augmented LLMs can serve as powerful tools for automating complex algorithmic optimizations that traditionally require deep domain expertise and extensive manual tuning. As the demand for efficient vector search continues to grow with the proliferation of RAG and agent-based LLM applications, automated optimization frameworks like CRINN will become increasingly valuable for maintaining competitive performance across evolving hardware architectures and application requirements.

\bibliography{custom}
\bibliographystyle{acm}
\end{document}

%% file: images/overall_results_plot.tex
\begin{figure*}[h]
 \centering
 \begin{adjustbox}{margin=-0.8cm 0cm 0cm 0cm}
 \begin{minipage}[c]{\textwidth}
 \centering
\includegraphics[scale=0.23]{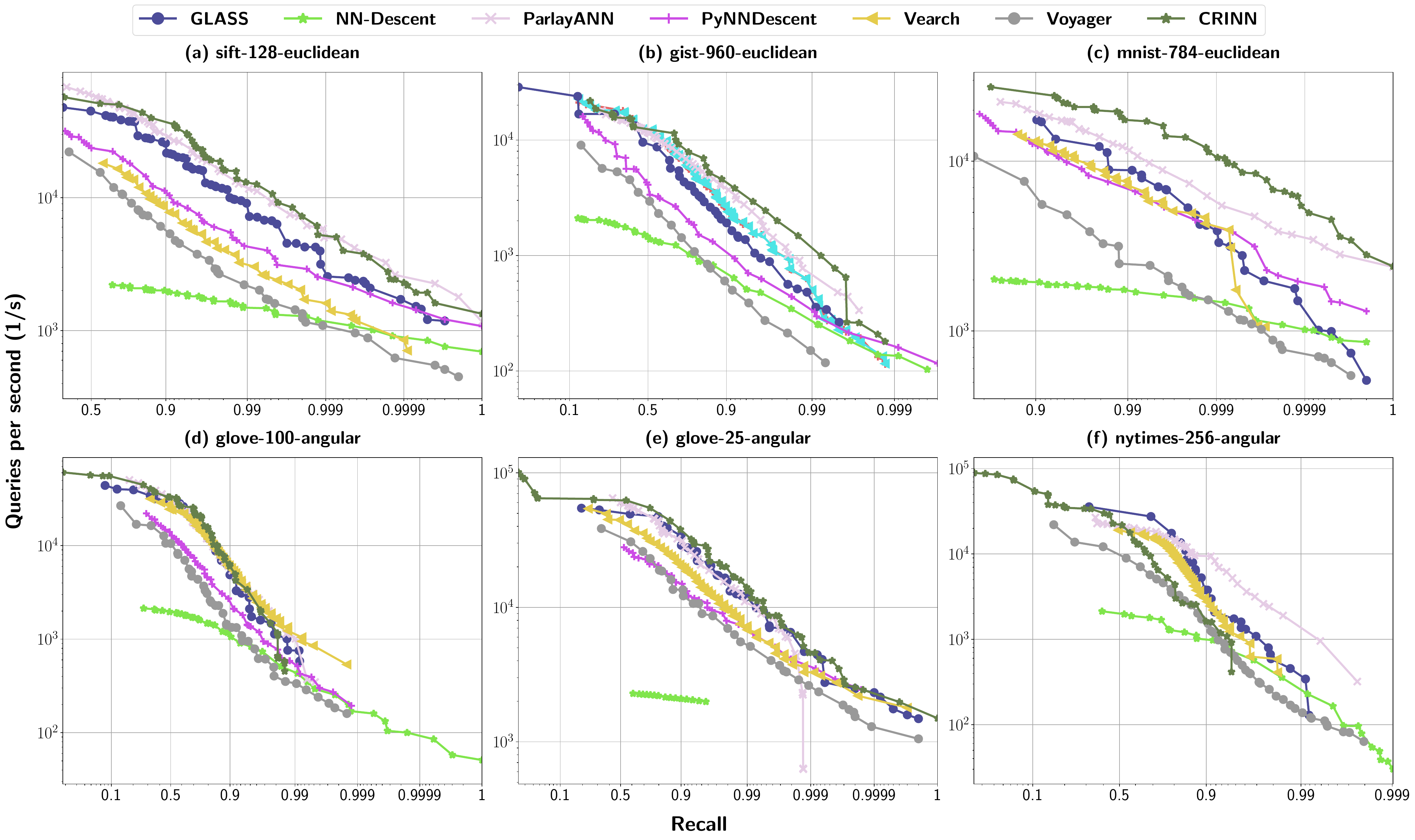}
\caption{QPS versus recall curves for different models across six datasets. 
CRINN achieves
achieves best-in-class performance on three out of them (GIST-960-Euclidean, MNIST-784-Euclidean, and GloVe-25-angular) and matching state-of-the-art results on two (SIFT-128 and GloVe-25). }
\label{fig:overall_result_plot}
 \end{minipage}
 \end{adjustbox}
 \end{figure*}

%% file: tables/rl_prompt_template.tex
\begin{table}[htbp]
  \centering

  \begin{tcolorbox}[
    rounded corners,
    arc=3pt,
    enhanced,
    width=0.96\textwidth,
    colframe=codegreen,
    colback=white,
    title=\textbf{Prompt Template Used in CRINN},
    fonttitle=\bfseries\large,
    boxrule=0.6pt,
    left=6pt,
    right=6pt,
    top=6pt,
    bottom=6pt
  ]
    \begin{tcolorbox}[
      sharp corners,
      colframe=taskblue,
      colback=taskblue!5,
      colbacktitle=taskblue!15,
      coltitle=taskblue,
      boxrule=0.4pt,
      title={\textbf{Task Description}},
      fonttitle=\bfseries,
      left=4pt,
      right=4pt,
      top=4pt,
      bottom=4pt,
      breakable
    ]
You are an approximate nearest neighbor search optimization expert specializing in high-performance similarity search algorithms. Given reference implementations for search, your objective is to create an accelerated version that maintains identical functionality. You will receive previous module implementations accompanied by their scores indicating the general speed. Higher scores indicate higher speed. Conduct a comparative analysis of these implementations and use the insights to develop optimized graph construction code.
    \end{tcolorbox}

    \begin{tcolorbox}[
      sharp corners,
      colframe=refgreen,
      colback=refgreen!5,
      colbacktitle=refgreen!15,
      coltitle=refgreen,
      boxrule=0.4pt,
      title={\textbf{Previous Implementations with Speed}},
      fonttitle=\bfseries,
      left=4pt,
      right=4pt,
      top=1pt,
      bottom=1pt,
      breakable,
      listing options={style=codestyle}
    ]
\begin{lstlisting}[language=C++, basicstyle=\small\ttfamily, style=codestyle, frame=none,framerule=0pt]
// Implementation 1 (Score: 1.42)
class Module_v1 {
    void build_index(const float* data, int n, int d) {
        ...
    }
    void search(const float* query, int k, int* indices, float* distances) {
        ...
    }
};
// Implementation 2 (Score: 1.34)
class Module_v2 {
    void build_index(const float* data, int n, int d) {
        ...
    }
    void search(const float* query, int k, int* indices, float* distances) {
        ...
    }
};
\end{lstlisting}
    \end{tcolorbox}

    \begin{tcolorbox}[
      sharp corners,
      colframe=subviolet,
      colback=subviolet!6,
      colbacktitle=subviolet!15,
      coltitle=subviolet,
      boxrule=0.4pt,
      title={\textbf{Generation Protocol}},
      fonttitle=\bfseries,
      left=4pt,
      right=4pt,
      top=4pt,
      bottom=4pt,
      breakable
    ]
You MUST use exactly two hash symbols (\#\#) at the beginning of each section. \\
 \textbf{\texttt{\#\# Performance Analysis}}: Compare ANNS implementations above and articulate on:
\begin{enumerate}[noitemsep]
\item Which implementations achieve superior query throughput and what algorithmic factors contribute to their fast execution?
\item What indexing structures or search strategies demonstrate the best speed-accuracy tradeoffs?
\item What are the primary bottlenecks limiting query performance in slower implementations?
\item Which vectorization, parallelization, or caching techniques remain unexploited?
\end{enumerate}
\vspace{0.3mm}
\textbf{\texttt{\#\# Algorithm Design}}: Describe your optimization strategy as numbered points outlining key techniques and improvements for accelerating execution speed  \\
\textbf{\texttt{\#\# Code}}: Your code implementation 
\end{tcolorbox}

    \begin{tcolorbox}[
      sharp corners,
      colframe=orange!90!black,
      colback=orange!6,
      colbacktitle=orange!15,
      coltitle=orange,
      boxrule=0.4pt,
      title={\textbf{Requirements and Constraints}},
      fonttitle=\bfseries,
      left=4pt,
      right=4pt,
      top=4pt,
      bottom=4pt,
      breakable
    ]
 \textbf{\texttt{\#\# Critical Requirements}}: 
\begin{enumerate}[noitemsep]
\item Search quality must match the reference implementation exactly (same recall, precision). Failure to maintain search accuracy will result in a score of 0.
\item The module must support the same interface: build\_index() and search() methods with identical parameters.
\item Results must be deterministic and reproducible across runs.
\end{enumerate}

\end{tcolorbox}

  \end{tcolorbox}
  \caption{Prompt template used in CRINN for RL training.}
\label{tab:prompt_example}
\end{table}

%% file: tables/recall.tex
\begin{table}[h]
\centering
\setlength{\tabcolsep}{5pt}
\renewcommand\arraystretch{1.1}
\begin{tabular}{@{}llrrrr@{}}
\toprule
\textbf{Dataset} & \textbf{Recall} & \textbf{CRINN} & \textbf{Best Baseline} & \textbf{Baseline} & \textbf{Improvement} \\
 & & \textbf{QPS} & & \textbf{QPS} & \\
\midrule
\multicolumn{6}{c}{\textit{Euclidean Distance}} \\[2pt]
SIFT-128 & 0.900 & 36,876 & ParlayANN & 29,368 & +25.57\% \\
         & 0.950 & 27,499 & ParlayANN & 23,057 & +19.26\% \\
         & 0.990 & 13,014 & ParlayANN & 11,808 & +10.21\% \\
         & 0.999 & 5,158  & ParlayANN & 4,996  & +3.25\% \\[3pt]
GIST-960 & 0.900 & 4,288  & ParlayANN & 3,788  & +13.20\% \\
         & 0.950 & 2,925  & ParlayANN & 2,348  & +24.59\% \\
         & 0.990 & 1,149  & ParlayANN & 666    & +72.68\% \\[3pt]
MNIST-784 & 0.900 & 24,826 & ParlayANN & 19,324 & +28.47\% \\
          & 0.950 & 22,008 & ParlayANN & 17,293 & +27.26\% \\
          & 0.990 & 17,457 & ParlayANN & 11,728 & +48.85\% \\
          & 0.999 & 10,600 & ParlayANN & 5,722  & +85.25\% \\[4pt]
          \midrule
\multicolumn{6}{c}{\textit{Angular Distance}} \\[2pt]
GloVe-100 & 0.900 & 5,947  & Vearch    & 5,768  & +3.09\% \\
          & 0.950 & 3,024  & ParlayANN & 3,212  & {-5.84\%} \\[3pt]
GloVe-25  & 0.900 & 37,474 & Glass     & 31,611 & +18.55\% \\
          & 0.950 & 28,909 & Glass     & 21,899 & +32.01\% \\
          & 0.990 & 13,574 & Glass     & 11,804 & +14.99\% \\
          & 0.999 & 4,588  & Glass     & 4,549  & +0.87\% \\[3pt]
NYTimes-256 & 0.900 & 1,623 & ParlayANN & 9,459  &{-82.85\%} \\
\bottomrule
\end{tabular}
\caption{Performance comparison of CRINN against best baselines across different datasets and recall levels. QPS (Queries Per Second) measures throughput.}
\label{tab:performance}
\end{table}

%% file: tables/progressive.tex
\begin{table}[h]
\centering
\setlength{\tabcolsep}{6pt}
\renewcommand\arraystretch{1.1}
\begin{tabular}{l|cc|cc|cc}
\toprule
\multirow{2}{*}{\textbf{Dataset}} & \multicolumn{2}{c|}{\textbf{Graph Construction}} & \multicolumn{2}{c|}{\textbf{Search}} & \multicolumn{2}{c}{\textbf{Refinement}} \\
& Individual & Cumulative & Individual & Cumulative & Individual & Cumulative \\
\midrule
sift-128-euclidean & +30.12\% & +30.12\% & +25.86\% & +55.98\% & +11.19\% & +67.17\% \\
gist-960-euclidean & +58.26\% & +58.26\% & +46.43\% & +104.69\% & +29.63\% & +134.32\% \\
mnist-784-euclidean & +45.85\% & +45.85\% & +44.49\% & +90.34\% & +18.30\% & +108.64\% \\
glove-100-angular & +13.19\% & +13.19\% & +19.03\% & +32.22\% & +5.86\% & +38.08\% \\
glove-25-angular & +6.94\% & +6.94\% & +6.52\% & +13.46\% & +2.70\% & +16.16\% \\
nytimes-256-angular & -21.68\% & -21.68\% & -32.54\% & -54.22\% & -9.56\% & -63.78\% \\
\midrule
\textbf{Overall Average} & \textbf{+22.11\%} & \textbf{+22.11\%} & \textbf{+18.30\%} & \textbf{+40.41\%} & \textbf{+9.69\%} & \textbf{+50.10\%} \\
\bottomrule
\end{tabular}
\caption{Average QPS improvement across different recall levels.}
\label{table:progressive}
\end{table}